\pdfoutput=1

\documentclass[11pt]{article}

\usepackage[final]{acl}

\usepackage{times}
\usepackage{latexsym}

\usepackage[T1]{fontenc}

\usepackage[utf8]{inputenc}

\usepackage{microtype}

\usepackage{inconsolata}

\usepackage{graphicx}
\usepackage{xspace}
\usepackage{paralist}
\usepackage{cleveref}
\usepackage{multirow}
\usepackage{booktabs}

\usepackage{adjustbox}
\usepackage[utf8]{inputenc}
\usepackage[draft]{minted}

\usepackage{listings}
\usepackage{makecell}
\usepackage{seqsplit}
\usepackage{xcolor}
\usepackage{caption}
\usepackage{subcaption}
\usepackage{amssymb}
\usepackage{listings}
\usepackage[normalem]{ulem}
\usepackage{amssymb}
\usepackage{wasysym}

\definecolor{lightbg}{RGB}{242, 242, 242}
\definecolor{lighttext}{RGB}{30, 30, 30}
\definecolor{lightkeyword}{RGB}{0, 0, 255}
\definecolor{lightcomment}{RGB}{0, 128, 0}
\definecolor{lightstring}{RGB}{163, 21, 21}
\definecolor{lightnum}{RGB}{153, 0, 153}
\definecolor{lightbuiltin}{RGB}{0, 128, 128}
\definecolor{lightidentifier}{RGB}{0, 0, 0}
\definecolor{lightvariable}{RGB}{0, 0, 0}
\definecolor{lightfunction}{RGB}{128, 0, 128}

\lstdefinestyle{pythonstyle}{
    language=Python,
    backgroundcolor=\color{lightbg},
    basicstyle=\footnotesize\ttfamily\color{lighttext},
    keywordstyle=\color{lightkeyword},
    commentstyle=\color{lightcomment},
    stringstyle=\color{lightstring},
    numberstyle=\color{lightnum},
    identifierstyle=\color{lightidentifier},
    tabsize=2,
    showstringspaces=false,
    breaklines=true,
    frame=tb,
    framesep=4pt,
    numbersep=8pt,
    numberstyle=\tiny\color{gray},
    moredelim=[s][\color{lightbuiltin}]{@}{\ },
    morekeywords={self},
}

\definecolor{codegreen}{rgb}{0,0.6,0}
\definecolor{codegray}{rgb}{0.5,0.5,0.5}
\definecolor{codepurple}{rgb}{0.58,0,0.82}
\definecolor{backcolour}{rgb}{0.95,0.95,0.92}

\lstdefinestyle{mystyle}{
    backgroundcolor=\color{backcolour},   
    commentstyle=\color{codegreen},
    keywordstyle=\color{magenta},
    numberstyle=\tiny\color{codegray},
    stringstyle=\color{codepurple},
    basicstyle=\ttfamily\scriptsize,
    breakatwhitespace=false,         
    breaklines=true,                 
    captionpos=t,                    
    keepspaces=true,                 
    numbers=none, 
    numbersep=5pt,                  
    showspaces=false,                
    showstringspaces=false,
    showtabs=false,                  
    tabsize=2,
    showlines=true
}

\newcommand{\model}{PropTest\xspace}
\newcommand{\eg}{e.g.,\xspace}

\title{\model: Automatic Property Testing for Improved Visual Programming}

\author{Jaywon Koo $^\spadesuit$,  Ziyan Yang $^\spadesuit$, Paola Cascante-Bonilla $^\spadesuit$, \\
\textbf{Baishakhi Ray $^\blacklozenge$,  Vicente Ordóñez $^\spadesuit$}\\
Rice University $^\spadesuit$ \quad Columbia University $^\blacklozenge$ \\
  \texttt{\{jk125, zy47, pc51, vicenteor\}@rice.edu}\\
  \texttt{rayb@cs.columbia.edu} 
  }

\begin{document}
\maketitle

\begin{abstract}
Visual Programming has recently emerged as an alternative to end-to-end black-box visual reasoning models. This type of method leverages Large Language Models (LLMs) to generate the source code for an executable computer program that solves a given problem. This strategy has the advantage of offering an interpretable reasoning path and does not require finetuning a model with task-specific data.
We propose \model, a general strategy that improves visual programming by further using an LLM to generate code that tests for visual properties in an initial round of proposed solutions. Our method generates tests for data-type consistency, output syntax, and semantic properties. 
\model achieves comparable results to state-of-the-art methods while using publicly available LLMs.
This is demonstrated across different benchmarks on visual question answering and referring expression comprehension.
Particularly, \model improves ViperGPT by obtaining 46.1\% accuracy (+6.0\%) on GQA using Llama3-8B and 59.5\% (+8.1\%) on RefCOCO+ using CodeLlama-34B.
\end{abstract}

\section{Introduction}
\label{sec:intro}

Visual reasoning tasks often require multi-hop reasoning that goes beyond surface-level observations. This type or reasoning typically involves complex multi-step processes, external knowledge, or understanding of compositional relationships between objects or entities. 
End-to-end vision and language models based on deep neural networks trained with huge amounts of data are used to tackle these tasks~\citep{li2023blip2,alayrac2022flamingo,yu2022coca,driess2023palm,li2022grounded, wang2023cogvlm}. However, these methods often fail at multi-hop compositional reasoning as they aim to solve a wide array of reasoning tasks in a single forward pass. Recent work has proposed Visual Programming as a principled way to tackle visual reasoning~\citep{gao2023fine, suris2023vipergpt, gupta2022visual, subramanian-etal-2023-modular}. These techniques work by leveraging a Large Language Model (LLM) to generate the logic of a program in the form of its source code that can be used to solve the problem. These methods can combine various tools in complex ways and offer interpretability and the opportunity to diagnose failures in their predicted logic. 

\begin{figure*}[t!]
    \centering
    \includegraphics[width=0.98\textwidth]{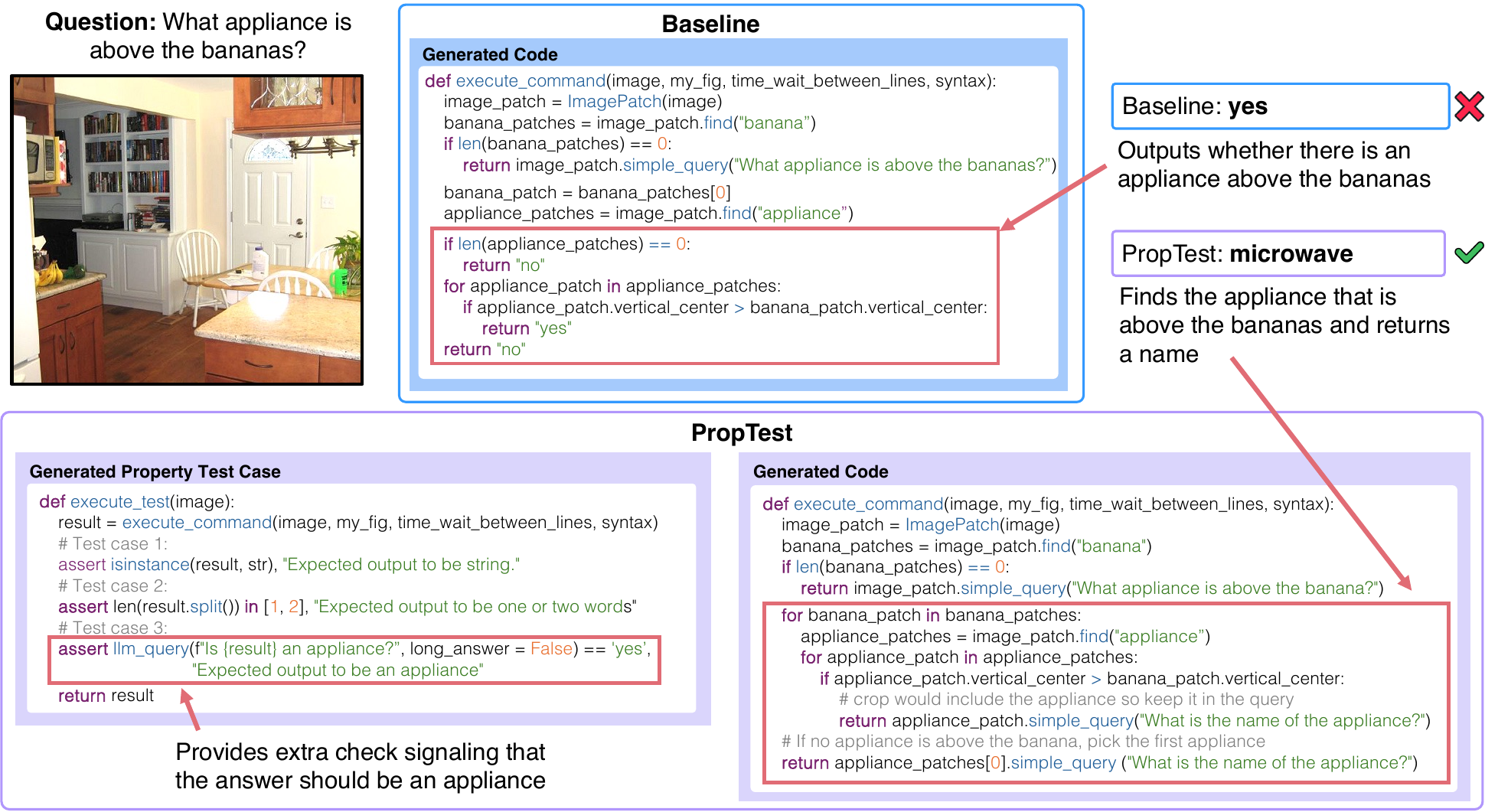}
    \vspace{-0.05in}
    \caption{Visual programming methods generate code for a program to solve a vision-and-language task such as VQA. PropTest improves on these methods by automatically generating testing code that probes for several output properties. This is used as additional information when generating code and checking the correctness of the output solutions. As a baseline we use ViperGPT under CodeLlama-7B for this example.}
    \label{fig:ViperOurs}
    \vspace{-0.15in}
\end{figure*}

Visual programming methods that rely on code generation and program execution to solve a task still rely on end-to-end pre-trained Vision Language Models (VLMs) either as tools that can be invoked by the program or as a {\em fallback} option when the generated code contains syntax or runtime errors. In other words, if the generated code contains errors, then a default end-to-end VLM is invoked. For these methods to be effective, the generated source code should produce solutions that lead to correct results on average more often than their {\em fallback} VLM. However, there are still many instances where a generated source code contains no syntax or runtime errors, but the logic of the program produces results that contain incorrect logic to solve the problem. Some of these are easier to spot, such as instances where the code returns the wrong data type, or the wrong type of answer for the given problem (e.g. answering with a location when the question is about a quantity). We posit that code testing and assertion error checking which are established practices in software development, should also help these types of methods in guiding them toward better solutions.

We introduce \model, a visual programming framework that generates automatic property test cases to guide code generation and identify logic that is likely to contain errors. 
Fig.~\ref{fig:ViperOurs} showcases a motivating example for our proposed method.  
\model first generates property test cases using an LLM which probes for data type inconsistencies, syntactic errors, and semantic properties of the results. 
For instance, in the showcased question {\it What appliance is above the bananas?}, the generated test code anticipates that the answer should be a Python {\texttt{string}} data type, that it should be limited to one or two words, and that the output should be a type of {\em appliance}. 
We find that this type of tests consistently help the LLM generate code for the program that is less likely to contain errors.  

\begin{figure*}[t]
    \centering
    \includegraphics[width=\textwidth]{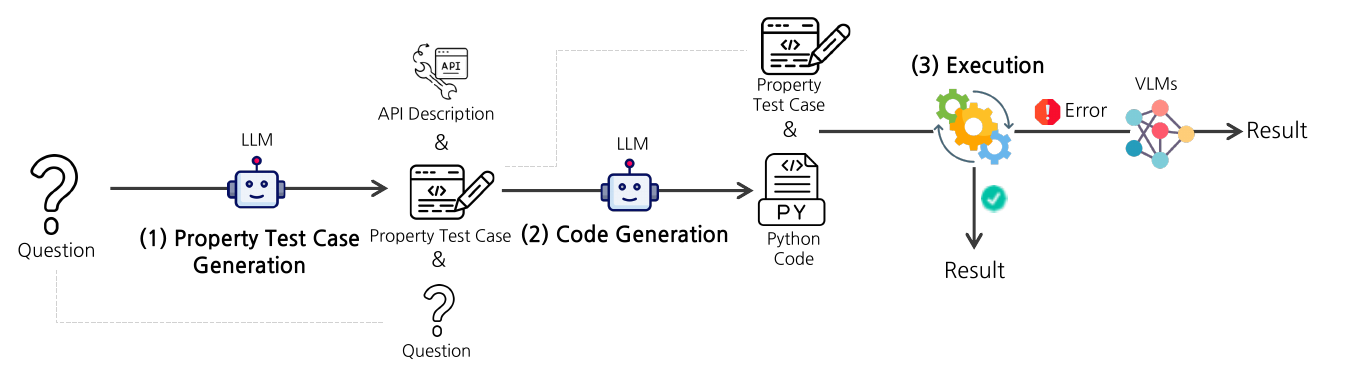}
    \caption{An overview of \model. Given an image and a question, the goal is to generate Python code that can be executed to get an answer. 
    \model first calls an LLM to generate test cases based on the inferred properties of the answer. Then, the generated test cases are used to improve the quality of Python code. 
    }
    \label{fig:mainfig}
    \vspace{-0.1in}
\end{figure*}

\model can filter out incorrect outputs resulting from errors in logic or failures in dependent modules and redirect these cases when appropriate to the {\em fallback} VLM.
Moreover, \model provides additional information about failure cases and in characterizing the type of errors.
Additionally, previous visual programming methods rely on closed-source models, 
making it hard to reproduce results due to continuous version updates, deprecation of older models (\eg~Codex), and usage costs~\citep{gupta2022visual, suris2023vipergpt, subramanian-etal-2023-modular}. 
Our main experiments rely exclusively on public models, such as \textsc{CodeLlama}~\citep{roziere2023code} and \textsc{llama3}~\cite{llama3modelcard}, which we expect to serve as stable baselines for future work on this area. 
We evaluate \model on three different tasks: Compositional visual question answering (GQA~\citep{hudson2019gqa}), External knowledge-dependent image question answering (A-OKVQA~\citep{schwenk2022okvqa}), and Visual grounding (RefCOCO and RefCOCO+~\cite{yu2016modeling}). 
Our experiments show that property tests significantly enhance performance across these benchmarks.
We also analyze detailed errors from a software engineering perspective (assertion, runtime, and syntax).

Our contributions can be summarized as follows:
\begin{compactitem}
    \item We propose \model, a novel framework that uses automatic property test case generation for detecting logic, syntax, and runtime errors, which are used to guide code generation. 
    \item \model improves interpretability when errors occur, bridging the gap between LLMs and VLMs on code generation.
    \item Our proposed method obtains superior results on four benchmarks compared to a baseline model conditioned on four different publicly available LLMs and one proprietary LLM. 
\end{compactitem}

\section{Method}
We introduce \model, a framework for leveraging property test code generation. 
A commonly recommended practice in software development is to write tests first and then write the code for the logic of the program so that it passes the tests. 
This is the responsible programmer approach to software development. 
We emulate this approach in \model by first generating testing code and then generating code to solve the task conditioned on the testing code. 
Fig.~\ref{fig:mainfig} shows an overview of our method.

Let us consider a question such as {\it What kind of toy is the boy playing with?}, we can easily infer that the answer should be a type of {\em toy}. 
We utilize this insight to provide information to the code generation model, narrowing down the search space rather than only relying on single-step prompt optimization.
Additionally, generating property test cases is generally simpler than generating code because test cases are shorter and more straightforward. 
Creating an easier test case first sets a baseline to generate more complex code. 
Property test cases guide the code generation process and increase the likelihood of generating accurate and effective code solutions.

Our framework first generates property test cases using an LLM by providing a problem statement as a prompt, \eg~a question, or a referring expression.
The source code for these generated tests is then added to the prompt of the LLM, along with the original problem statement and detailed API documentation of the available tools or modules.
We employ the same API and tools used in ViperGPT~\citep{suris2023vipergpt}, which also relies on generic functions from the Python programming language. The code generation model then outputs the code solution that addresses the problem statement and returns a plausible result.

\begin{figure*}[t!]
    \centering
    \includegraphics[width=0.98\textwidth]{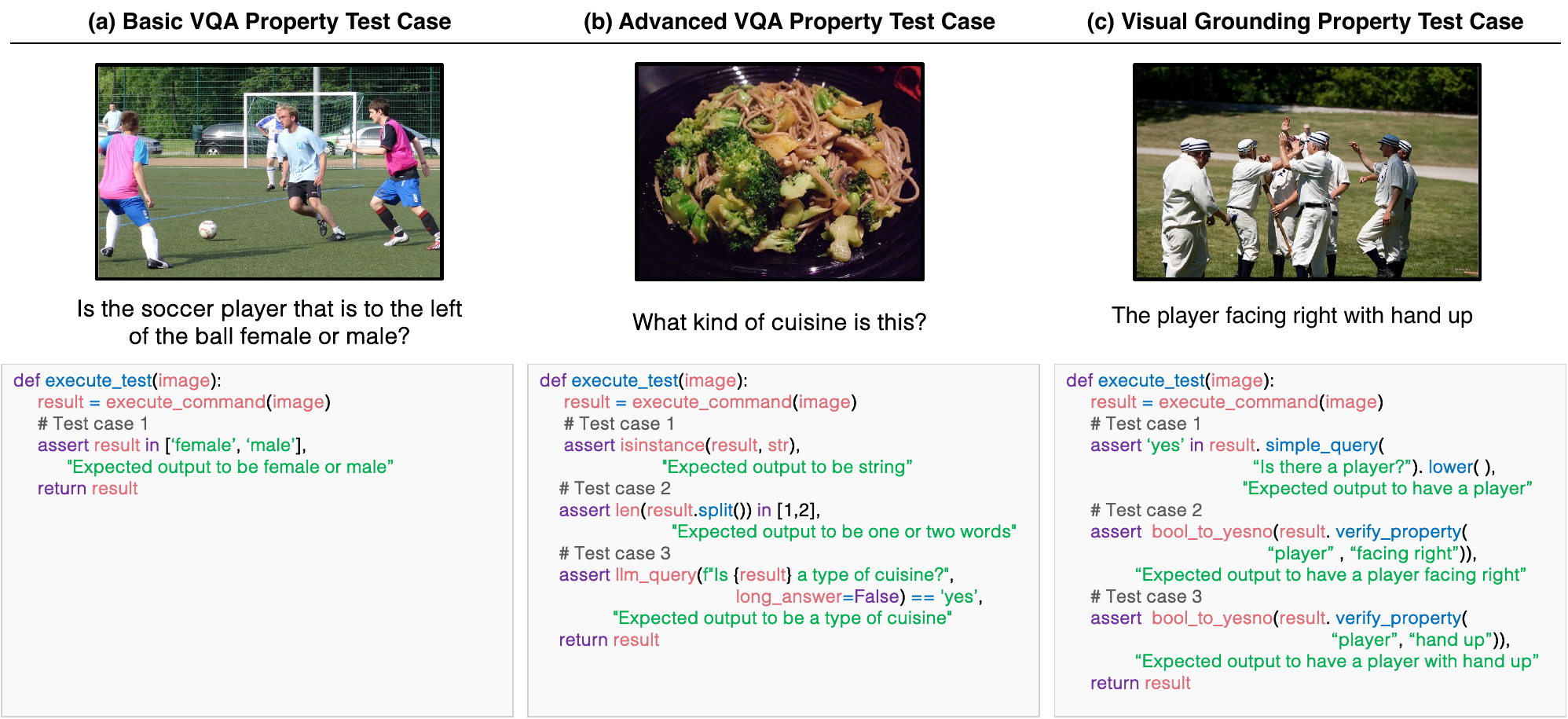}
    \vspace{-0.1in}
    \caption{Three different examples of property test cases generated for visual question answering and for visual grounding. The \texttt{execute\_command()} is the generic name of the generated program code routine and \texttt{result} is the output from executing it.}
    \label{fig:testcase}
    \vspace{-0.15in}
\end{figure*}

We concatenate the generated property test case and the code solution and apply an execution engine where we also provide the visual input. 
There can be a syntax or runtime error inside the generated main code.
An assertion error will occur if the code output does not pass any of the property test cases.
If execution proceeds without errors, including syntax, runtime, or assertion errors, the result is returned, and the process concludes.
In the event of an error, we 
default to a task-specific {\em fallback} VLM and return. 

\section{Property Test Case Generation}

The purpose of using a property tests is to verify whether a generated code works as expected. 
Our property tests guide an LLM to generate better code that meets basic properties.
The design of property test cases varies based on the data type of the answer due to the different tools (APIs) available for each type.
In this section, we explain in detail the design process for prompts used to generate property tests for visual question answering tasks, where the task answer is text~(\cref{sec:ImgQATask}) and for visual grounding tasks, where the task answer is an image with bounding boxes~(\cref{sec:ImgGroundTaks}).

\subsection{Property Tests for Visual Question Answering}
\label{sec:ImgQATask}
Visual question answering tasks contain queries that require multi-hop reasoning or external knowledge. 
To solve these tasks, we propose two property test case generation strategies along with corresponding in-context prompts to guide the LLM toward the generation of property tests with similar logic. We include our prompts in~\Cref{sec:appendix:prompt}.

\vspace{0.02in}
\noindent
\textbf{Basic Property Test Case Generation.}
This type of test only relies on basic Python functions without using external APIs or tools.  
As shown in Fig.~\ref{fig:testcase}a, this approach is effective when the question mentions several candidates.  
Furthermore, this strategy can be applied to yes-or-no questions, where it checks the type of the property. 

\vspace{0.02in}
\noindent
\textbf{Advanced Property Test Case Generation.}
For this type of test cases, we also allow the use of tools through an API specification, specifically the use of an LLM that can check the output result through various properties. Particularly, our generated test code can use an \texttt{llm\_query()} function to construct more advanced assertion statements. Fig.~\ref{fig:testcase}b shows an example where given the question {\it What kind of cuisine is this?}, 
the first test case checks the return data type, which should be a Python \texttt{string}. 
Then a second assertion checks that the output is just one or two words in length. The third test case checks the semantic property of the returned result. 
Knowing that the expected answer should be a type of {\em cuisine}, we use LLM queries in the test case to verify whether the result correctly identifies a {\em cuisine} type.
This effectively narrows the expected result space for the code generation model, helping it produce more accurate solutions. 

\subsection{Property Tests for Visual Grounding}
\label{sec:ImgGroundTaks} 
Visual grounding tasks require returning a bounding box in an image that corresponds to an input text query. 
To construct property test cases for such tasks, we utilize a set of tools that take images as inputs.
Particularly, our test code can use functions such as \texttt{simple\_query()}, \texttt{verify\_property()}, and \texttt{bool\_to\_yesno()}.
The \texttt{simple\_query()} function is used to answer straightforward questions about the image, \texttt{verify\_property()} checks whether an object has a given attribute as a property, and \texttt{bool\_to\_yesno()} converts boolean values into "yes" or "no" responses. 
As shown in Fig.~\ref{fig:testcase}c, given the input referring expression {\it the player facing right with hand up}, our test case begins by confirming if a player is inside the result bounding box.
It then proceeds to verify, in sequence, whether the identified player is facing {\it right} with {\it hand up}, thus checking whether the given output is likely to reflect the given query.

\begin{figure*}[ht]
    \centering
     \begin{subfigure}[b]{0.48\textwidth}
        \centering
        \includegraphics[width=\textwidth]{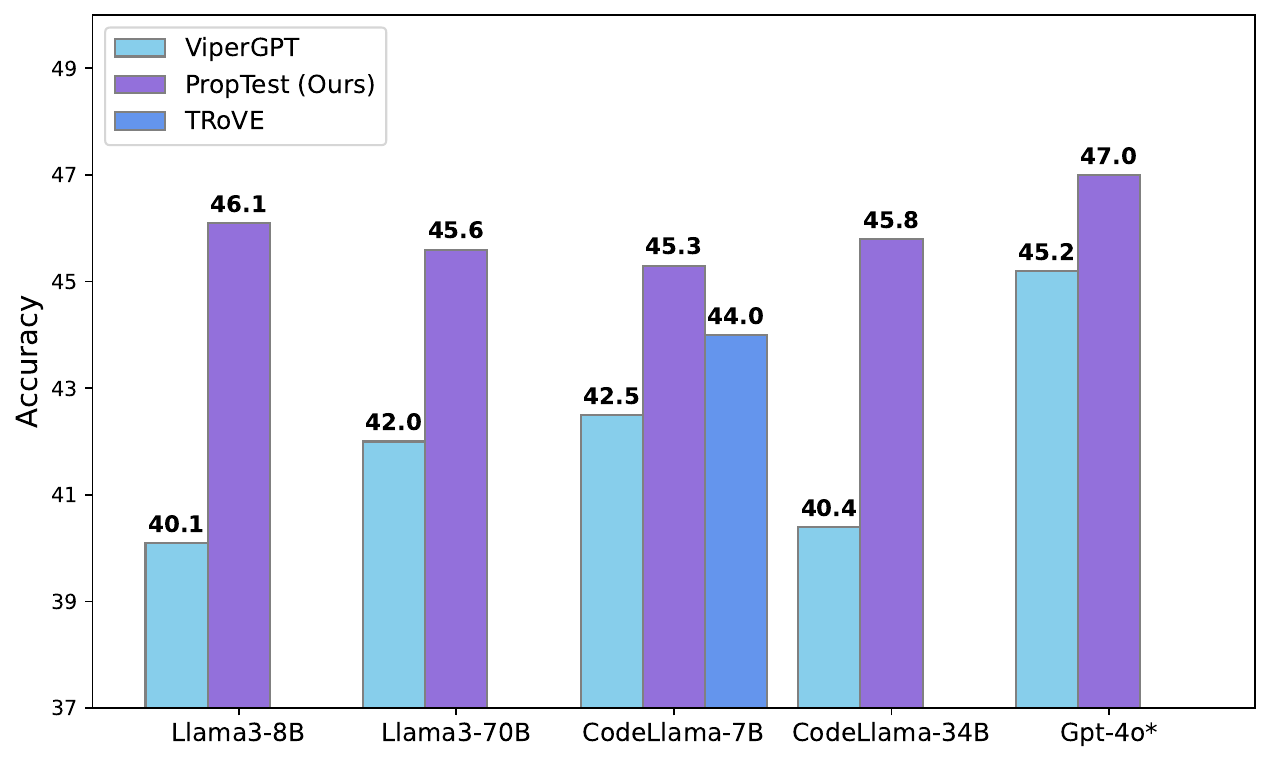}
        \caption{Results on GQA using different LLMs}
        \label{fig:gqa}
    \end{subfigure}
    \hfill
    \begin{subfigure}[b]{0.48\textwidth}
        \centering
        \includegraphics[width=\textwidth]{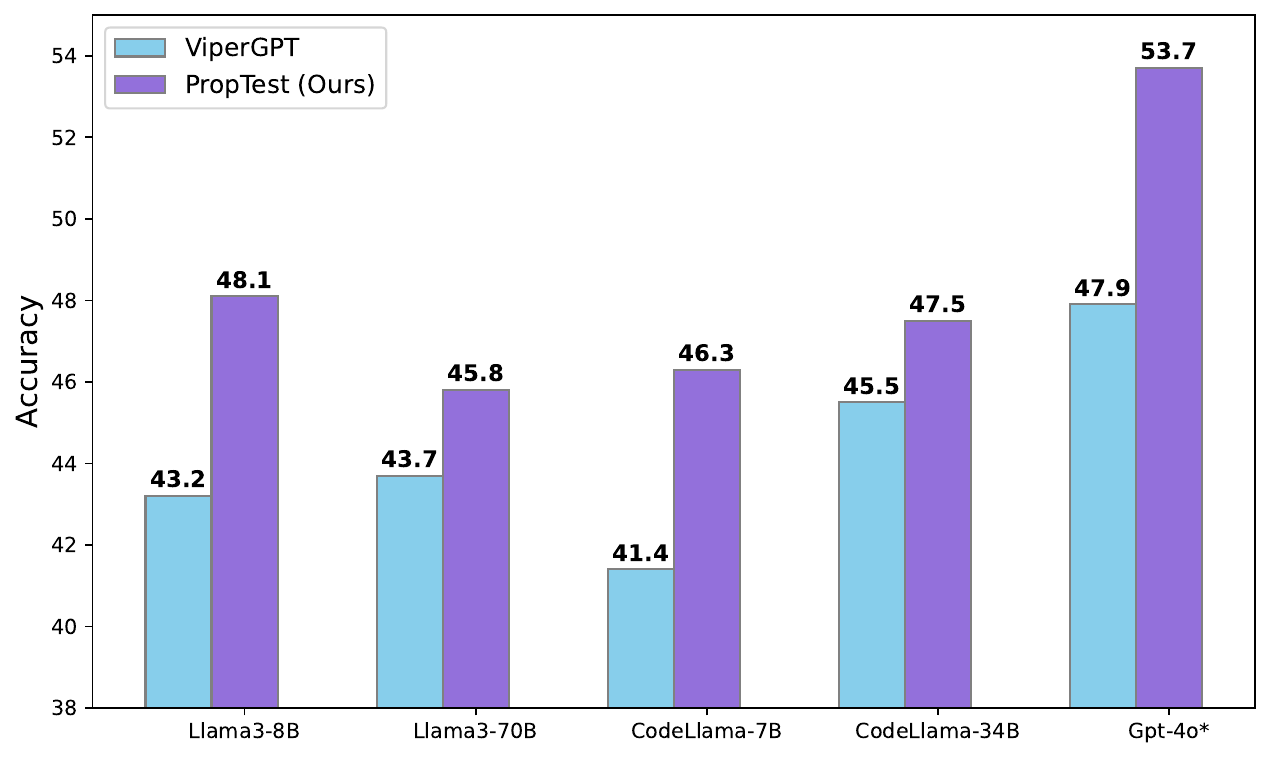}
        \caption{Results on A-OKVQA using different LLMs}
        \label{fig:aokvqa}
    \end{subfigure}
    \vfill
    \begin{subfigure}[b]{0.48\textwidth}
        \centering
        \includegraphics[width=\textwidth]{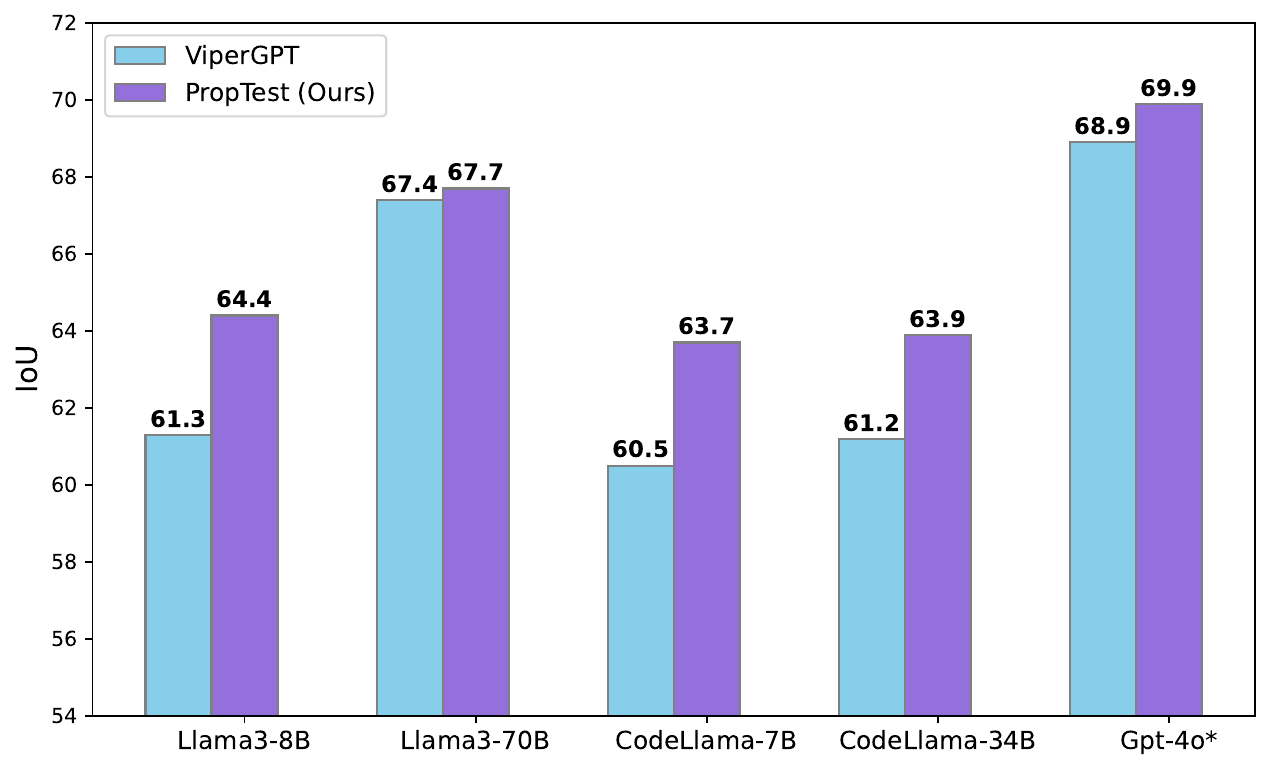}
        \caption{Results on RefCOCO using different LLMs}
        \label{fig:Refcoco}
    \end{subfigure}
    \hfill
    \begin{subfigure}[b]{0.48\textwidth}
        \centering
        \includegraphics[width=\textwidth]{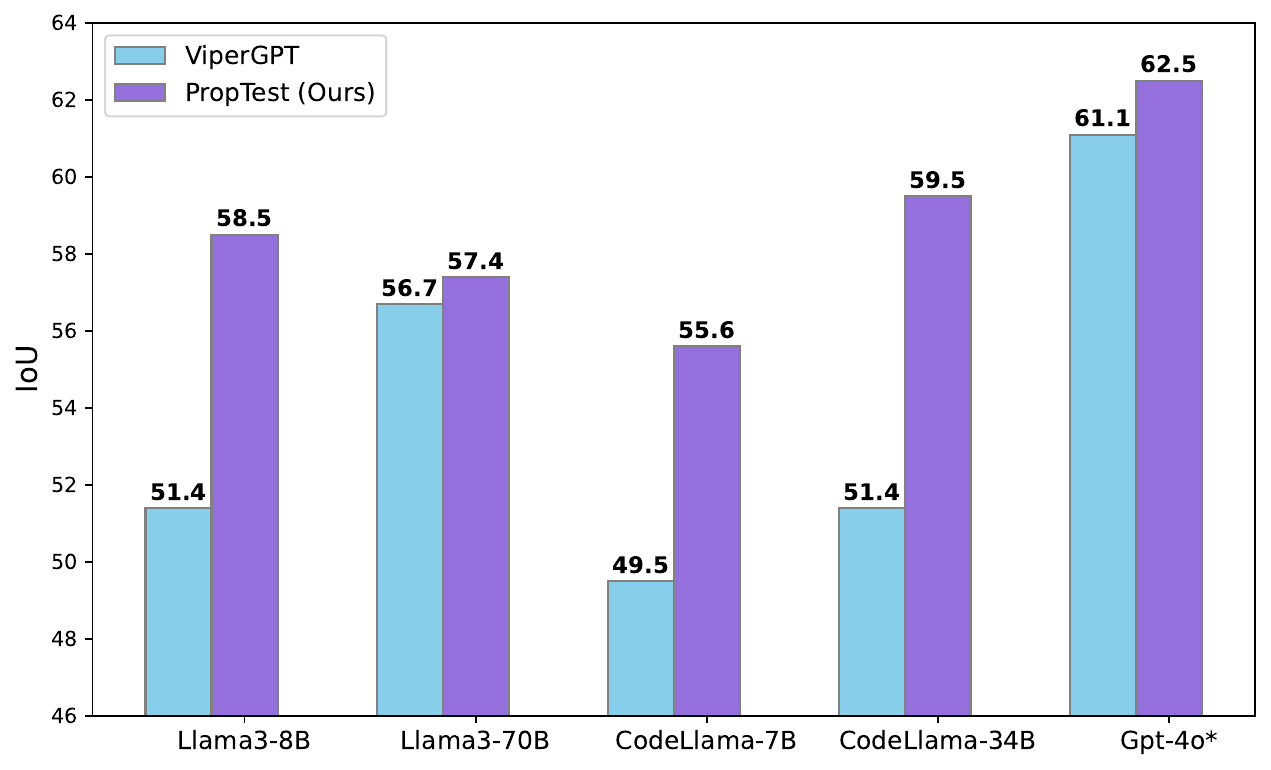}
        \caption{Results on RefCOCO+ using different LLMs}
        \label{fig:Refcoco+}
    \end{subfigure}
    \caption{Comparison of our method against visual programming methods with different LLMs across two tasks, four benchmarks. We report Accuracy on two visual question answering benchmarks, and IoU on two visual grounding benchmarks. GPT-4o* results are only tested on 500 subsamples.}
    \label{fig:quantitative_Results}
    \vspace{-0.15in}
\end{figure*}

\section{Experiments} 
We introduce the experimental setup~(\cref{sec:4.1ExpSetup}), and report the results on different LLMs~(\cref{sec:4.2results})

\subsection{Experimental Setup}
\label{sec:4.1ExpSetup}
\noindent
\textbf{Tasks and Metrics.} We validate \model on the Visual Question
Answering (VQA) and Visual Grounding tasks. For VQA, we evaluate on GQA~\cite{hudson2019gqa}, and A-OKVQA~\cite{schwenk2022okvqa}, which contain complex multi-hop questions that require compositional reasoning skills. 
We adopt exact match accuracy as our evaluation metric for GQA, where answers must correspond with a single ground truth answer.
We use soft accuracy~(SAcc)~\cite{antol2015vqa} for A-OKVQA. For Visual Grounding, we choose standard benchmarks, including testA split on RefCOCO and RefCOCO+~\cite{yu2016modeling}. 
The evaluation metric is the intersection over union (IoU) score. 

\noindent
\textbf{Model Comparison.} Similar to prior work, for VQA we use BLIP-2~\cite{li2023blip2} as our {\em fallback} VLM, and GLIP~\cite{li2022grounded} for Visual Grounding. 
The tools and API specifications for \model are consistent with those employed by ViperGPT~\cite{suris2023vipergpt}, ensuring a standardized basis for comparison. 
Therefore, for our experimental comparisons, we compare \model with other code generation models - ViperGPT~\cite{suris2023vipergpt}, and end-to-end models including BLIP-2~\cite{li2023blip2} and GLIP~\cite{li2022grounded}. 
The only other publicly available neuro-symbolic method is the concurrent work from~\citet{wang2024trove}, which uses \textsc{CodeLLama-7B}.

\noindent
\textbf{Implementation Details.}
We implement \model using the open-source LLMs including \textsc{CodeLlama} (\textsc{7B}, \textsc{34B})~\cite{roziere2023code} and \textsc{Llama3 (8B, 70B)}~\cite{llama3modelcard} for code generation. 
The specific implementation details are described in~\Cref{sec:appendix_experimental}. 

\begin{figure*}[t!]
    \centering
    \includegraphics[width=0.98\textwidth]{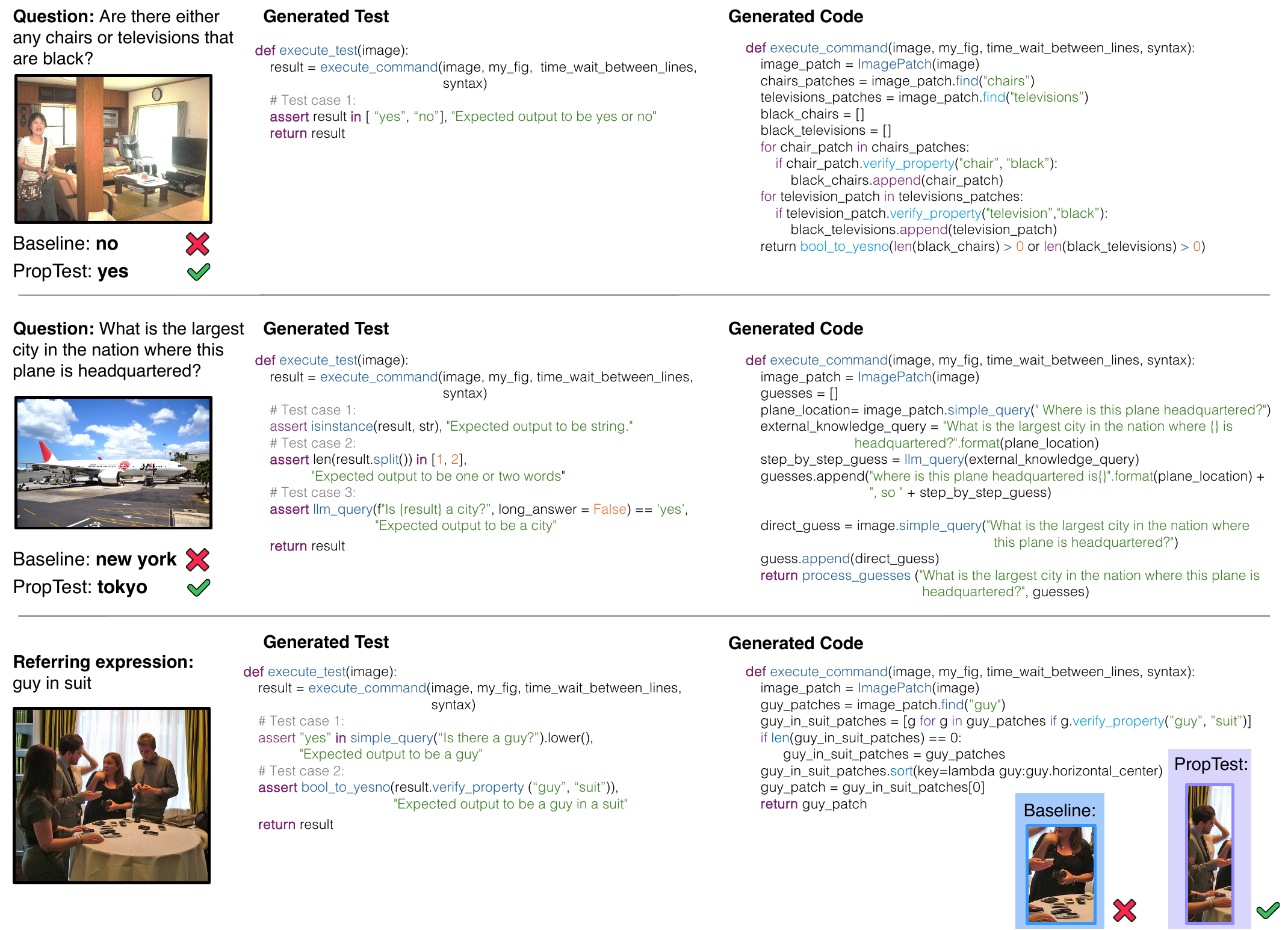}
    \vspace{-0.1in}
    \caption{Example results on GQA, A-OKVQA and RefCOCO. We show cases where PropTest succeeds but the baseline ViperGPT fails. Input questions and answers are shown on the left, generated property test cases in the middle, and code on the right.}
    \label{fig:qualitativeResult}
    \vspace{-0.2in}
\end{figure*}
\subsection{Results}
\label{sec:4.2results}
\textbf{Quantitative Results.}
One common concern with previous work is that evaluations performed with API-based black-box models (e.g.~GPT-3.5, GPT-4) are hard to reproduce and track as there are many different upgrades on these models.
They can also be discontinued (e.g.~Codex), making past work non-reproducible. Our main experiments are conducted using \textsc{CodeLlama} and \textsc{Llama3}, which are publicly available and free to use for research purposes. As part of our work, we will also release an API-free implementation of ViperGPT. 
Additionally, we evaluate \model using GPT-4o to contextualize our work. We limit our evaluation to 500 randomly sampled subsets for each data split, specifically for GPT-4o.

Our main results are shown in Fig.~\ref{fig:quantitative_Results}. 
Overall, \model shows improvements over ViperGPT in all settings. 
The model that provides the most gain varies by dataset, smaller models such as CodeLlama-7B and Llama3-8B tend to benefit more with PropTest (\eg +6.0\% on GQA with Llama3-8B, +4.9\% on A-OKVQA with both LLMs and +7.1\% on RefCOCO+ with Llama3-8B) but even larger models also show gains, including GPT-4o. 
Notably, CodeLlama-34B outperforms or shows greater improvement over ViperGPT compared to Llama3-70B across all datasets. This is due to CodeLlama-34B's training with code, making it superior in code generation despite its smaller size relative to Llama3-70B.
We also noticed that GPT-4o shows the best results on all datasets.

Moreover, \model outperforms the {\em fallback} VLMs we rely on, while also providing enhanced interpretability in all settings.
The {\em fallback} VLM results are 42.4\%\footnote{Result under the same setting as ViperGPT, differing from the original work~\cite{li2023blip2}} on GQA, 45.1\% on A-OKVQA, 55.0\% on RefCOCO, and 52.2\% on RefCOCO+. 
While ViperGPT sometimes underperforms compared to VLMs depending on the LLMs, \model remains robust, performing well on all models, including smaller ones. 

We did not compare our models to previous visual programming methods that use closed API-based LLMs~\citep{yuan2023craft,subramanian-etal-2023-modular,gnsvr2023}, as it would be unfair or unfeasible due to the different or deprecated LLMs used in those models.

\noindent
\textbf{Qualitative Results.} 
Fig.~\ref{fig:qualitativeResult} shows representative examples of the types of property tests that get generated and output programs. 
By leveraging property test cases, \model generates a code with correct logic and results on cases that fail to return a correct answer due to logical errors on ViperGPT. 
In addition, we illustrate cases with logical errors that produce assertion errors in~\Cref{sec:appendix_error}. By checking on logical errors, \model provides extra interpretability on the reason for failure.
More qualitative results are shown in~\Cref{sec:appendix_qualitative}.

\section{Error Analysis \& Discussion}

In this section, we first focus on the question: \textit{What types of errors does the code generation model produce?} 
We analyze the errors in the generated code from ViperGPT and \model across datasets, categorizing them into three basic Python errors: Assertion, Runtime, and Syntax errors.
We report results using Llama3-8B in~\Cref{table:error_analysis}.
\begin{table}[t]
\centering
\small
\setlength\tabcolsep{1.5pt}
\renewcommand{\arraystretch}{1.1}
\begin{tabular}{llcccc}
\toprule
\textbf{Dataset}   & \textbf{Method}        & \textbf{\# Errors}  & \textbf{Assert.} & \textbf{Runt.} & \textbf{Syntax} \\ \midrule
\multirow{ 2}{*}{GQA}       & ViperGPT & 411 (3.3\%)        & -         & 322     & 89     \\
          & PropTest                   & 1264 (10.0\%)      & 1001      & 227     & 36     \\ \midrule
\multirow{ 2}{*}{A-OKVQA}   & ViperGPT & 11 (1.0\%)         & -         & 9       & 2      \\
          & PropTest                    & 174 (15.2\%)       & 169       & 3       & 2      \\ \midrule
\multirow{ 2}{*}{RefCOCO}   & ViperGPT & 281 (5.0\%)        & -         & 240     & 41     \\
          & PropTest                    & 871 (15.4\%)       & 617       & 241     & 13     \\ \midrule
\multirow{ 2}{*}{RefCOCO+}  & ViperGPT & 435 (7.6\%)        & -         & 386     & 49     \\
          & PropTest                    & 1132 (19.8\%)      & 875       & 250     & 7      \\ \bottomrule
\end{tabular}%
\caption{Error Analysis on ViperGPT~\cite{suris2023vipergpt} and PropTest across benchmarks using Llama3-8B including runtime and syntax errors.}
\label{table:error_analysis}
\vspace{-0.2in}
\end{table}

We first note that code generation models produce more errors in visual grounding tasks than in VQA tasks. This is because visual grounding involves stricter assertions in test cases, leading to a higher frequency of assertion errors. In visual grounding, all test cases check the result \texttt{image\_patch} for specific properties, and errors occur when objects or properties are missing. In contrast, VQA often involves simpler yes-or-no checks, where incorrect results might still pass the test.
Furthermore, RefCOCO+ has a higher overall error rate compared to RefCOCO due to its complex queries. The simpler queries in RefCOCO make \model generate test cases that accurately identify the target object, resulting in fewer errors. Detailed analysis with examples is in~\Cref{sec:appendix_error}.

We also find that due to additional assertion errors, \model has higher overall errors compared to ViperGPT. 
Nevertheless, \model notably reduces runtime and syntax errors on three datasets (\eg $322 \rightarrow 227$ runtime, $89 \rightarrow 39$ syntax errors in GQA). This reduction indicates that the inclusion of property test cases enhances code generation quality in the aspects of runtime and syntax errors. 
However, the increase in assertion errors, leading to a rise in total errors, implies that \model relies more on the {\em fallback} model. This raises the question: \textit{Does the performance gain of \model come from an increased dependence on VLMs?}

To address this, we compare the performance of ViperGPT and \model without using the \textit{fallback} model for error handling, as shown in \Cref{tab:ablation_VLMs}. Across all datasets, \model either outperforms or performs on par with ViperGPT, demonstrating that the performance gain is from improved code quality rather than increased reliance on VLMs. 

\begin{table}[t]
\centering
\small
\renewcommand{\arraystretch}{1.1}
\setlength\tabcolsep{1.8pt}
\begin{tabular}{lccccc}
\toprule
& \multicolumn{2}{c}{w/o VLMs as fallback} && \multicolumn{2}{c}{w/ VLMs as fallback} \\

\cmidrule{2-3}\cmidrule{5-6}

\textbf{Dataset} & \textbf{\scriptsize ViperGPT} & \textbf{\scriptsize \model} &&  \makecell{\textbf{\scriptsize \model} \\ \textit{\scriptsize w/o running tests}} & \textbf{\scriptsize \model}  \\
\midrule
GQA       & 39.1     & 43.8     && 45.8 & 46.1 \\
A-OKVQA   & 42.8     & 42.8     && 47.3 & 48.1 \\
RefCOCO   & 60.1     & 61.6     && 63.8 & 64.4 \\
RefCOCO+  & 50.2     & 55.8     && 58.1 & 58.5 \\
\bottomrule
\end{tabular}

\caption{
Ablation study on the reliance on Visual Language Models (VLMs) for error handling in generated code and the impact of executing test cases.
}
\label{tab:ablation_VLMs}
\vspace{-0.1in}
\end{table} 
Now, we move on to another question: \textit{How does running a test case during execution help when there is an error?} To address this, we compare \model with an approach that does not run test cases when errors occur. Our findings show that running test cases in the presence of errors increases accuracy, indicating that our generated property test cases are effective at detecting incorrect code (\eg +0.8 in A-OKVQA). 

\section{Property Test Analysis}
\label{sec:TestCase_analysis}
In this section, we investigate generated property tests in depth by comparing two types of VQA property test cases (section~\ref{sec:6.1Basic_vs_Advanced}) and evaluating the generated property test cases (section \ref{sec:6.2testcaseeval}).

\subsection{Basic vs Advanced Property Tests }
\label{sec:6.1Basic_vs_Advanced}

\Cref{tab:Basic_Advanced_Compare} shows the accuracy and error analysis of two types of VQA property test cases using Llama3-8B. 
Advanced property test cases have higher accuracy compared to basic tests. 
Using advanced property test case generation produces almost twice as many errors as basic property test case generation. This is due to an extra semantic property test, which leads to more assertion errors.
Advanced property test cases will be longer and more complicated than basic test cases, which causes more syntax errors (\eg $31 \rightarrow 36$).

\subsection{Generated Property Test Evaluation}
\label{sec:6.2testcaseeval}
We first evaluate our generated property tests on correctness by using the answers. 
If an answer passes the generated test, we count it as correct. 
We report this as accuracy in~\Cref{tab:testcase_acc}. 
We also examine the quality of our property test cases by using toxicity rate~\cite{chen2022codet}.
If the produced results pass the test while the answer fails the test, we assume the test case is {\em toxic}. 
Advanced VQA property test cases have lower accuracy and higher toxic rates compared to basic VQA tests because they generate complicated property test cases that check semantic properties using tools. 

\begin{table}[t!]
\centering
\small
\renewcommand{\arraystretch}{1.1}
\setlength\tabcolsep{2pt}
\begin{tabular}{lccccc}
\toprule
\textbf{Method} & \textbf{Acc.} & \textbf{\# Errors} & \textbf{Assert.} & \textbf{Runt.} & \textbf{Syntax} \\
\midrule
Basic VQA       & 45.6        & 732 (5.8\%)       & 469                & 232              & 31 \\
Advanced VQA    & 46.1        & 1264 (10\%)     & 1001               & 227              & 36 \\
\bottomrule
\end{tabular}
\caption{Error analysis on GQA dataset using basic and advanced property tests using Llama3-8B, including runtime and syntax errors. APIs are used for the Advanced VQA property test cases, where only basic Python functions are used in Basic VQA.}
\label{tab:Basic_Advanced_Compare}
\end{table}

\begin{figure}[t]
  \centering
  \includegraphics[width=0.65\columnwidth]{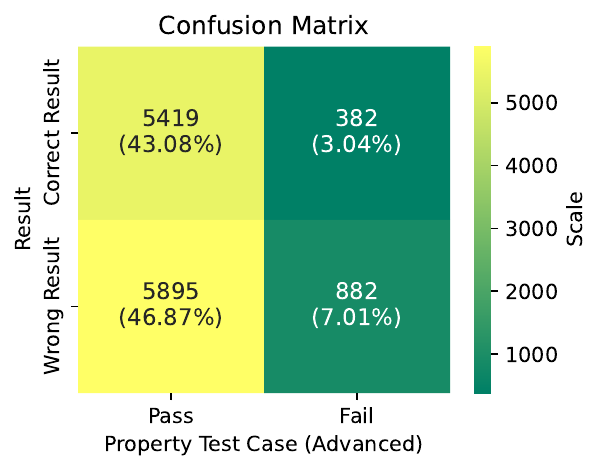}
  \vspace{-0.1in}
  \caption{Confusion Matrix of the generated advanced property test cases on GQA using Llama3-8B. We show the counts of correct and incorrect results, further divided by whether they passed or did not pass the generated property test case.}
  \label{fig:Advanced_confusion_matrix}
  \vspace{-0.2in}
\end{figure}

Moreover, we present a $2\times2$ confusion matrix for the advanced property test cases generated on GQA using Llama3-8B in Fig.~\ref{fig:Advanced_confusion_matrix}. The matrix shows a high number of false positives, primarily due to the flexibility of VQA property test cases. For example,  these tests often check for binary answers (yes or no), which can pass even if the result is incorrect. 
The confusion matrix for the basic property test case and for the visual grounding test case are provided in~\Cref{sec:appendix_genPropTest}. 
\begin{table}[t]
\centering
\small
\renewcommand{\arraystretch}{1.1}
\begin{tabular}{llcc}
\toprule
\textbf{Method} & \textbf{Dataset} & \textbf{Acc.} & \textbf{Toxic rate} \\
\midrule
Basic VQA      & GQA               & 95.7\%         & 0.03\% \\
Advanced VQA   & GQA               & 91.7\%         & 0.04\% \\
\bottomrule
\end{tabular}

\caption{Accuracy and toxic rate of generated property test cases on GQA with Llama3-8B. APIs are utilized in Advanced VQA property test cases, while only basic Python functions are used in Basic VQA.}
\label{tab:testcase_acc}
\vspace{-0.1in}
\end{table}

\section{Related Work}
\label{sec:Related_Work}
End-to-end vision language models (VLMs) are generally trained on large datasets containing images paired with text descriptions or instructions~\cite{li2023blip2,alayrac2022flamingo,yu2022coca,driess2023palm,li2022grounded, liu2023improved, guo2023images, wang2023cogvlm}. 
By learning correlations between visual features and linguistic patterns, VLMs can understand sophisticated relations between images and text using a single forward pass through a deep neural network. These models, however large, are still bounded by what functions can be learned and encoded in their model weights.

On the other hand, with the rise of LLMs for code generation in recent years~\cite{chen2021evaluating,roziere2023code,guo2024deepseek,nijkamp2023codegen,luo2023wizardcoder}, a recent set of methods in visual recognition have adopted the use of these models to solve visual tasks using a hybrid approach where VLMs and other computer vision models are used as tools by one of these code generation LLMs to generate a program that can solve a given task~\cite{suris2023vipergpt, gupta2022visual, subramanian-etal-2023-modular}. This type of neuro-symbolic reasoning model was referred to as {\em Visual Programming} by~\citet{gupta2022visual}. These methods lead to an executable program that decomposes complex visual reasoning queries into interpretable steps, which are then executed to produce results.
These methods define APIs (tools) they use during the execution, with functions mapped to off-the-shelf vision modules such as object detectors~\cite{he2017mask, li2022grounded}, depth estimators~\cite{ranftl2020towards}, among many others.   
These methods benefit from not needing extra training while enhancing reasoning capabilities and interpretability.
The performance of these methods depends on the tools or APIs the model leverages and the quality of the generated code. 
One line of work focuses on creating better and more diverse toolsets to improve accuracy~\cite{yuan2023craft, gnsvr2023, wang2024trove}.
Efforts to enhance code quality have been made by code refinement techniques, incorporating various types of feedback, such as visual, textual, error-related, and human feedback~\cite{gao2023fine}. Self-tuning mechanisms have also been explored to optimize model hyperparameters automatically~\cite{stanic2024towards}.
Our proposed method builds upon these findings, aiming to maximize the efficacy of VLMs~\cite{li2023blip2,li2022grounded} through property testing that is more specific to the visual domain.

Meanwhile, writing test cases is a common technique used by software developers to avoid writing code that contains programming errors. 
Similarly, it has enhanced code generation in code contest tasks. Test cases are used to detect errors and give feedback for self-refinement~\cite{le2023codechain, chen2023teaching, olausson2023self}.
Another line of work generates test cases by mutating existing test inputs~\cite{li2022competition} or by using LLMs~\cite{chen2022codet}.
Our research, however, differs from these methods by generating property test cases that check different properties of the output, and utilizing these test cases as an additional input when generating code. 

\section{Conclusion}
This paper presents \model, a novel framework for leveraging property test case generation to improve the quality of generated program code in visual programming.
\model shows consistent improvements on VQA and Visual Grounding datasets with four open-source code generation LLMs. 
Interestingly, we find that common software development advice which dictates that we should first write testing code before implementing new functionality, also applies to LLM-based code generation.

\newpage
\section{Limitations}
\model is an initial work that applies property test case generation for visual reasoning. Although the \model is a very promising framework for visual reasoning, there are several limitations that can be mentioned. 
First, \model requires an extra LLM inference to generate property test code, which will require extra time and resources, but we expect that as faster LLMs are supported in the future, this becomes less of an issue. Additionally, \model needs to design a specific property test case prompt depending on the type of the result (image or text). This can be resolved by adding an LLM that can design an automatic prompt depending on the task. 

Although less common, the code generated for the property tests themselves could also contain logical errors which limits their usefulness, and additionally, the tools they rely upon could also introduce errors. 
These limitations can be resolved by integrating visual programming works focused on tool generation~\cite{yuan2023craft, wang2024trove} or self-refining~\cite{gao2023fine, stanic2024towards} to enhance the code generation skills.
Finally, although the discussed datasets show strong performance, numerous visual reasoning tasks, such as video causal/temporal reasoning, remain to be explored in future research.

\bibliography{main}

\newpage
\newpage
\appendix
\label{sec:appendix}
\section{Experimental Details}
\label{sec:appendix_experimental}
We provide a detailed description of APIs (tools) used in \model in Section~\ref{sec:appendix:API}, LLMs in Section~\ref{sec:appendix:LLM} and prompts in Section~\ref{sec:appendix:prompt}.

\subsection{APIs (Pretrained Model) Details}
\label{sec:appendix:API}
Here, we specify the APIs (tools) we used:

$\diamond$ \textbf{\texttt{llm\_query(), process\_guess()}}: We use Llama3-8B-Instruct~\cite{llama3modelcard} and set the model to generate at most 256 tokens, temperature as 0.6 and top\_p as 0.9. 

$\diamond$ \textbf{\texttt{verify\_property()}}: We use open vocabulary object detector, GLIP~\cite{li2022grounded} is used. We used the same version used in ViperGPT~\cite{suris2023vipergpt}.

$\diamond$ \textbf{\texttt{best\_text\_match()}}: Image-text embedding model, X-VLM~\cite{pmlr-v162-zeng22c} fine-tuned version for retrieval on MSCOCO is used, which is also used in ViperGPT.

$\diamond$ \textbf{\texttt{simple\_query()}}: We use BLIP2~\cite{li2023blip2} with Flan-T5 XXL from its official repository. 

$\diamond$ \textbf{\texttt{compute\_depth()}}: The “DPT\_Large” version from the PyTorch hub4 of MiDaS~\cite{ranftl2020towards} was used.

$\diamond$ \textbf{\texttt{find()}}: We use MaskRCNN~\cite{he2017mask} for detecting objects and GLIP for detecting people.

\subsection{LLM Details}
\label{sec:appendix:LLM}
\begin{table}[tphb]
\centering
\resizebox{\columnwidth}{!}{%
\begin{tabular}{l|l}
\toprule
\textbf{LLM}   & \textbf{Specific Model} \\ \midrule
Llama3-8B      & meta-llama/Meta-Llama-3-8B-Instruct   \\    
Llama3-70B     & meta-llama/Meta-Llama-3-70B-Instruct\  \\
CodeLlama-7B   & meta-llama/CodeLlama-7b-Instruct-hf   \\
CodeLlama-34B  & meta-llama/CodeLlama-34b-Instruct-hf  \\ 
Gpt-4o         & gpt-4o-2024-05-13           \\ \bottomrule
\end{tabular}%
}
\caption{Specific details of the LLMs we use in \model. We used Huggingface versions for public LLMs.}
\vspace{-0.1in}
\label{table:appendix_LLMs}
\end{table}
\Cref{table:appendix_LLMs} shows the specific models used for property test case and code generation. We set the temperature as 0 and top\_p as 1 to avoid randomness for all LLMs.

\subsection{Prompt Details}
\label{sec:appendix:prompt}
In this section, we provide prompts of \model. 
First, the \texttt{system prompt} we used for property test case generation is as follows: 
\begin{lstlisting}[language=Python, linewidth=\columnwidth, xleftmargin=0pt, xrightmargin=0pt]
You are an expert programming assistant. Only answer with a 
function starting with def execute_test.
\end{lstlisting}
For the code generation, we used the following \texttt{system prompt}:
\begin{lstlisting}[style=mystyle]
Only answer with a function starting def execute_command.
\end{lstlisting}

We used two different prompt templates for test case generation and two different prompt templates for code generation.
Fig.~\ref{fig:first_test} shows the first prompt template for property test case generation, used for GQA. Fig.~\ref{fig:Second_test} illustrates the second prompt template, which was used for property test case generation in A-OKVQA, RefCOCO, and RefCOCO+. For RefCOCO and RefCOCO+, we only used the first line of the guideline. 

The first prompt template for code generation, as depicted in Fig.~\ref{fig:First_code}, is applied to both GQA and A-OKVQA datasets. The API descriptions and in-context examples are derived from ViperGPT~\cite{suris2023vipergpt} but have been shortened for brevity. We also employed the same set of 8 in-context examples. For A-OKVQA, only the first two guideline points were used.

\begin{figure}[H]
\centering
\begin{lstlisting}[language=Python]
# CONTEXT #
The 'solve_query' function is a Python function that takes an image as input and returns an answer to a <<QUERY>> in a string format.

# OBJECTIVE #
Create a Python function named `execute_test` that checks the correctness of the `solve_query` function using the given <<QUERY>>.
<<EXAMPLES>> are the in-context examples.
Include up to four test cases, each with the comment `# Test case n:` above the assert statement, starting from 1.
Consider these guidelines when creating the test cases:
1. Keep in mind that the return values do not contain numbers.
2. If the Query is True or False questions, the return values will be yes or no.
3. If the Query gives options using "or", the return values will be one of the options.
4. Use the llm_query function to answer informational questions not concerning the image.

# STYLE #
technical, in a correct Python format

# TONE #
clear, precise, professional

# AUDIENCE #
Developers and engineers who will use the test functions to verify the correctness of the solve_query function

# RESPONSE #
Provide the function that start with 'def execute_test(image)' without any explanation.
Each test case should be commented with `#Test case n:` where `n` represents the test case number.

###
Here are some <<EXAMPLES>>:
{{{{{{ TEN IN-CONTEXT EXAMPLES GOES HERE  }}}}}} 
###
# Instruction #
Generate the the function execute_test for the following query:
<<Query>>: INSERT_QUERY_HERE
\end{lstlisting}
\vspace{-0.1in}
\caption{First prompt template used to generate a property test case. In-context examples are omitted for brevity.}
\label{fig:first_test}
\end{figure}
\begin{figure}[H]
\centering
\begin{lstlisting}[language=Python, xleftmargin=.0\textwidth, xrightmargin=.0\textwidth]
Q: Your task is to write a function using Python containing tests up to four to check the correctness of a solve_query function that solves a provided answer to the query.
You must write the comment "#Test case n:" on a separate line directly above each assert statement,
where n represents the test case number, starting from 1 and increasing by one for each subsequent test case.

Here are some examples:

<<<<< TEN IN-CONTEXT EXAMPLES >>>>>

Consider the following guidelines:
- Only answer with a function starting with def execute_test.
- Return value of the solve_query function is a string with one or two words.
- Use the llm_query function to answer informational questions not concerning the image.

Query: INSERT_QUERY_HERE
\end{lstlisting}
\vspace{-0.1in}
\caption{Second prompt template used to generate a property test case. This template was used for A-OKVQA, RefCOCO, and RefCOCO+. In-context examples are omitted for brevity.}
\label{fig:Second_test}
\vspace{-0.2in}
\end{figure}
\begin{figure}[H]
\centering
\begin{lstlisting}[language=Python]
from PIL import Image
from vision_functions import find_in_image, simple_qa, verify_property, best_text_match

<<<<< API DESCRIPTIONS >>>>>

# Examples of using ImagePatch

<<<<< 8 IN-CONTEXT EXAMPLES >>>>>

Write a function using Python and the ImagePatch class (above) that could be executed to provide an answer to the query.

Consider the following guidelines:
- Use base Python (comparison, sorting) for basic logical operations, left/right/up/down, math, etc.
- Assertion tests (below) is used to verify the expected output. Consider these when writing the function.
- Do not return None or "Unknown". If the answer is not found, return image_patch.simple_query("INSERT_QUERY_HERE") to ask a question about the image.

Query: INSERT_QUERY_HERE
Assertion tests:
INSERT_ASSERTION_TESTS_HERE
\end{lstlisting}

\vspace{-0.1in}
\caption{First prompt template used to generate a code. This template is used for GQA and A-OKVQA.  API descriptions and in-context examples are omitted for brevity.}
\label{fig:First_code}
\vspace{-0.1in}
\end{figure}

Fig.~\ref{fig:Second_code} depicts the second template for code generation, used for RefCOCO and RefCOCO+. The API descriptions are from ViperGPT, and in-context examples differ by dataset. Also, for RefCOCO+, we used the following guidelines:
\begin{lstlisting}
Consider these guidelines when creating the function:
 - Use base Python (comparison, sorting) for basic logical operations, left/right/up/down, math, etc.
 - Consider the properties of the expected returned `ImagePatch` object from the << ASSERTION_TESTS >> to write the function.
 - The function must only return an `ImagePatch` object. Do not return None.
 - If the object in the query is not found directly, attempt to find a person and check if the person possesses or is associated with the specified object (e.g., wearing specific clothing).

\end{lstlisting}

\begin{figure}[H]
\centering
\begin{lstlisting}[language=Python]
# Context #
We are working on a visual grounding task, which involves identifying and returning the specific area of an image that corresponds to a given << QUERY >>. Using the << IMAGE_PATCH_CLASS >>, we aim to generate a Python function named `execute_command` to solve this task.

<< IMAGE_PATCH_CLASS >>

{{{{{ API DESCRIPTIONS }}}}}

#####################

# Objective #
Write a function named `execute_command` using Python and << IMAGE_PATCH_CLASS >> to answer the given << QUERY >>. Use the provided << ASSERTION_TESTS >> to understand the expected properties of the `ImagePatch` object that the function should return.
Consider these guidelines when creating the function:
- Use base Python (comparison, sorting) for basic logical operations, left/right/up/down, math, etc.
- Consider the properties of the expected returned `ImagePatch` object from the << ASSERTION_TESTS >> to write the function.
- The function must only return an `ImagePatch` object. Do not return None.

Here are some <<EXAMPLES>>:

{{{{{ 11 IN-CONTEXT EXAMPLES }}}}}

#####################

# RESPONSE #
Provide the function that starts with 'def execute_command(image)' without any explanation.

#####################
# START GENERATING CODE #
Generate the the function 'execute_command' for the following << QUERY >> and << ASSERTION_TESTS >>.
<< QUERY >>: INSERT_QUERY_HERE
<< ASSERTION_TESTS >>:
INSERT_ASSERTION_TESTS_HERE
\end{lstlisting}
\vspace{-0.1in}
\caption{Second prompt template used to generate a code. This template is used for RefCOCO and RefCOCO+. API descriptions and in-context examples are omitted for brevity.}
\label{fig:Second_code}
\end{figure}

\section{Qualitative Results}
\label{sec:appendix_qualitative}
\begin{figure*}[htbp]
    \centering
    \includegraphics[width=0.98\textwidth]{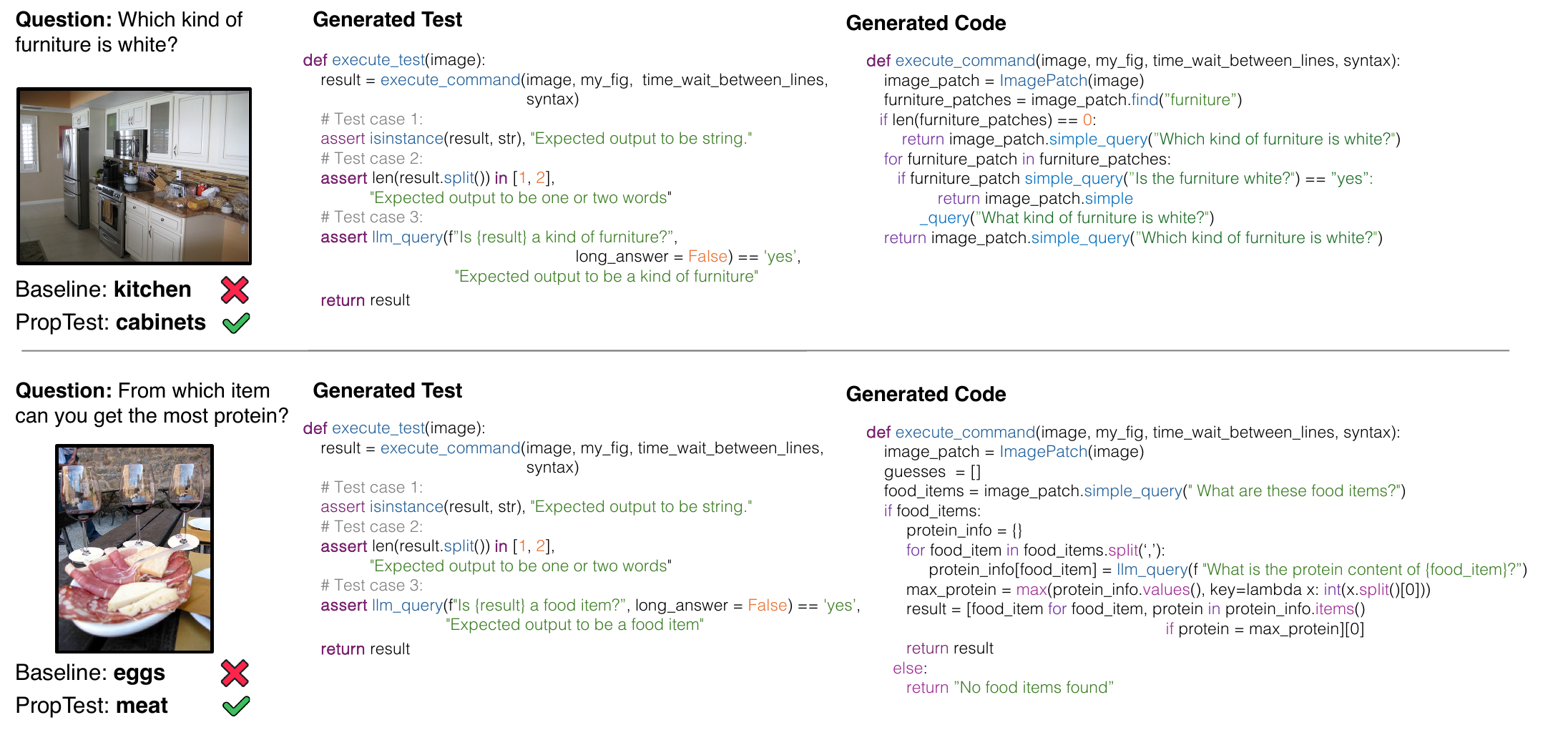}
    \caption{Example results on GQA and A-OKVQA. We present instances where PropTest is successful, whereas the baseline does not achieve the desired outcome. Input question and answer is shown on the left, generated property test case in the middle, code on the right and result on the left bottom.}
    \label{fig:Appendix_VQA_ex}
\end{figure*}
\begin{figure*}[htbp]
    \centering
    \includegraphics[width=0.98\textwidth]{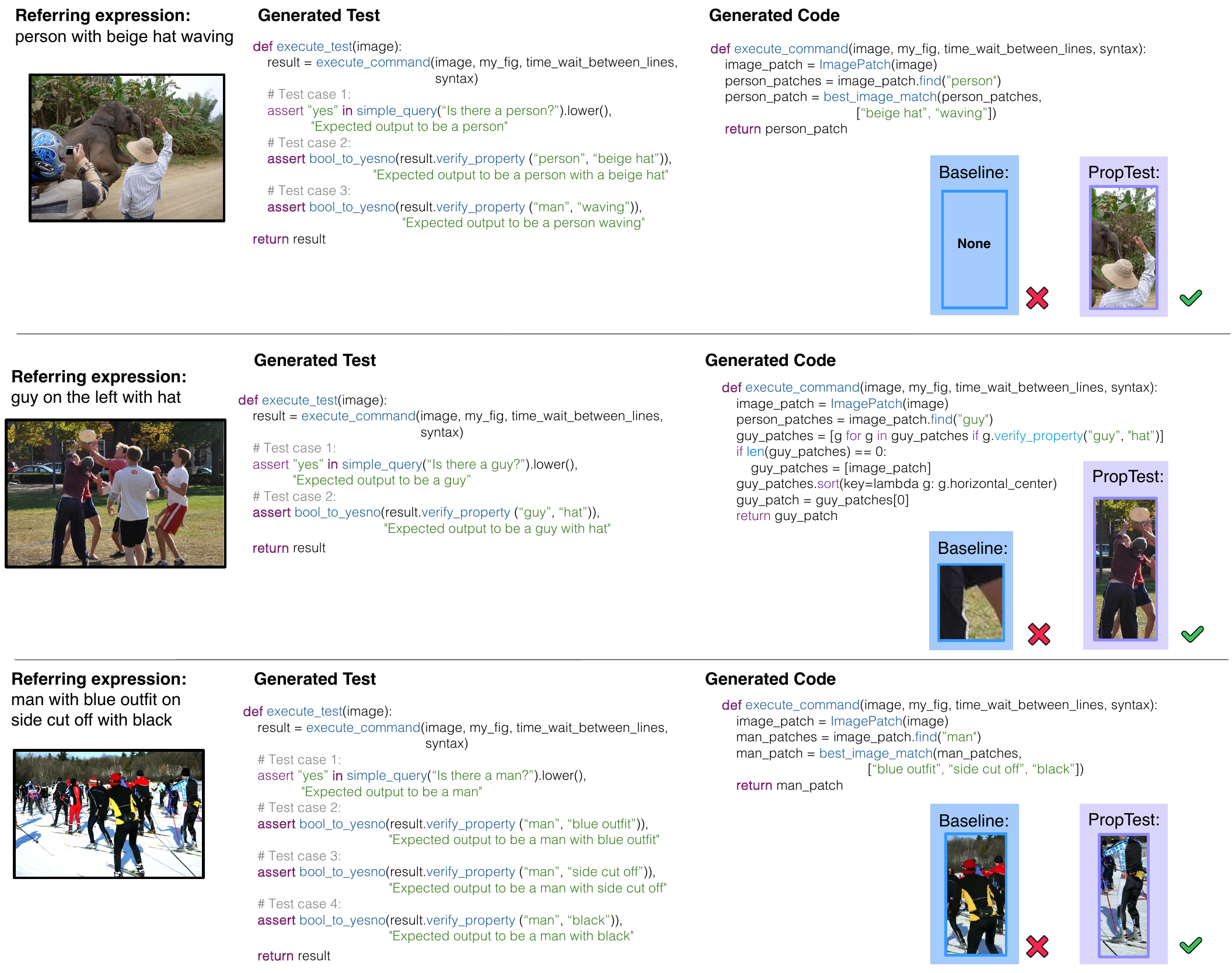}
    \caption{Example results on RefCOCO and RefCOCO+. We present instances where PropTest is successful, whereas the baseline does not achieve the desired outcome. Input question and answer is shown on the left, generated property test case in the middle, code on the right and result on the right bottom.}
    \label{fig:Appendix_VG_ex}
\end{figure*}
We provide additional examples across datasets. Fig.\ref{fig:Appendix_VQA_ex} plots the results on GQA and A-OKVQA and Fig.\ref{fig:Appendix_VG_ex} shows results on RefCOCO and RefCOCO+.
\section{Error Analysis}
\label{sec:appendix_error}
\begin{figure*}[tphb]
    \centering
    \includegraphics[width=\textwidth]{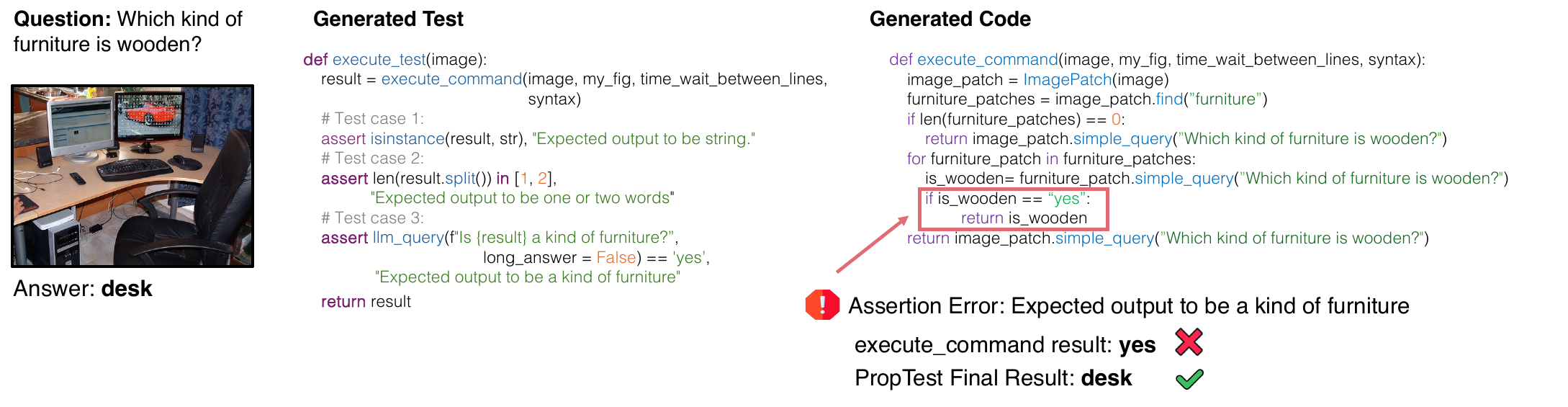}
    \caption{Example of failure case on GQA dataset using Llama3-8B where PropTest raises an assertion error. The final result is produced by BLIP-2~\cite{li2023blip2}. \model provides extra interpretability on the reason for failure by producing assertion errors.}
    \label{fig:appendix_assertionError}
\end{figure*}

We conduct a deeper analysis of the errors generated when using Llama3-8B. Fig.~\ref{fig:appendix_assertionError} shows a case where a property test case detects a logical error by raising an assertion error. By checking the properties of the result, \model identifies codes with incorrect logic and offers additional explanations for the failure.

\begin{figure*}[t]
    \centering
    \includegraphics[width=\textwidth]{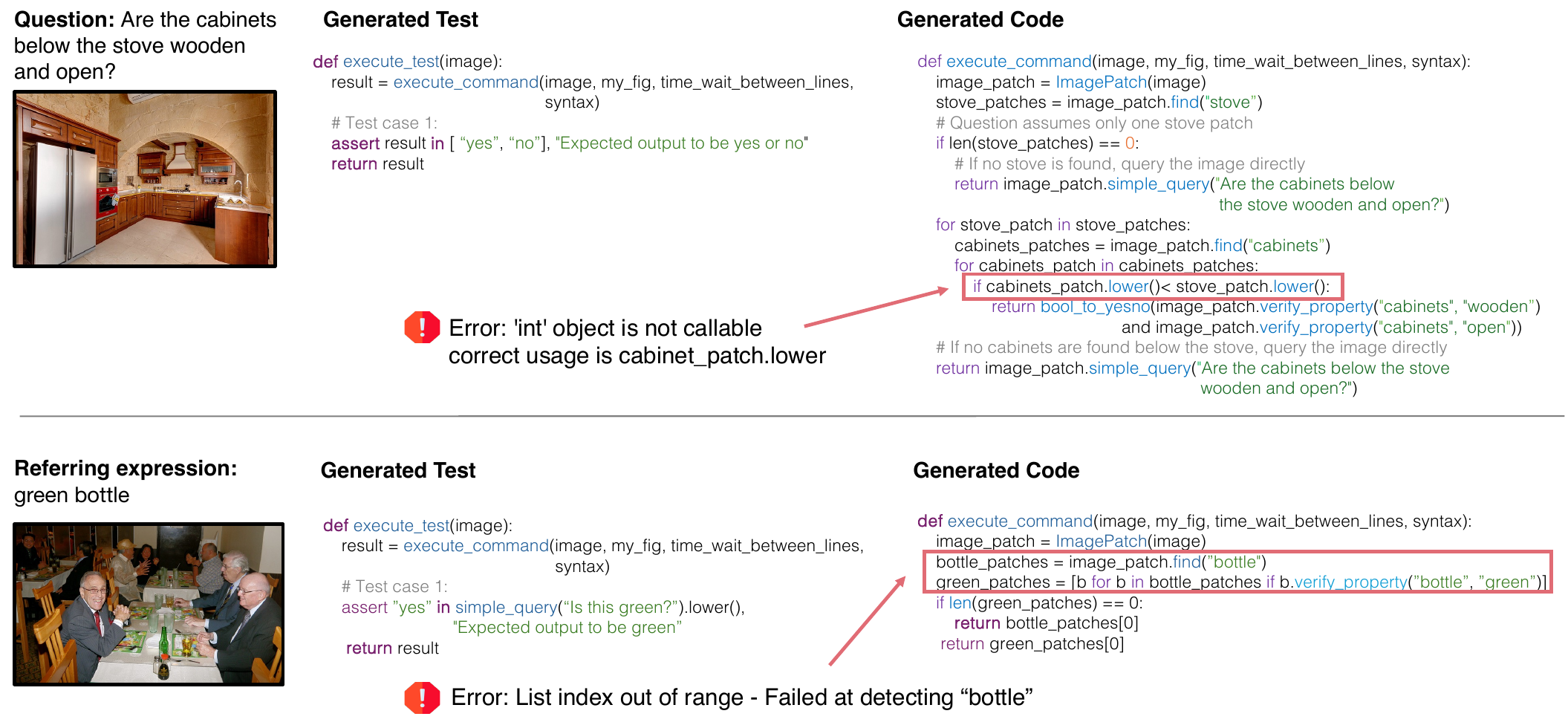}
    \caption{Examples of failure cases on GQA and RefCOCO dataset using Llama3-8B where PropTest raises a runtime error. \model provides extra interpretability on the reason for failure by producing assertion errors.
    }
    \label{fig:appendix_RuntimeError}
\end{figure*}

A number of runtime errors were detected across datasets. In GQA, the most common runtime error was due to incorrect usage of the attributes of \texttt{Class ImagePatch}, as shown in Fig.~\ref{fig:appendix_RuntimeError} (top). RefCOCO frequently encountered \texttt{List index out of range} errors, caused by the failure of the tool \texttt{find()} to detect an object (Fig.~\ref{fig:appendix_RuntimeError} (bottom)).

\begin{figure*}[t]
    \centering
    \includegraphics[width=\textwidth]{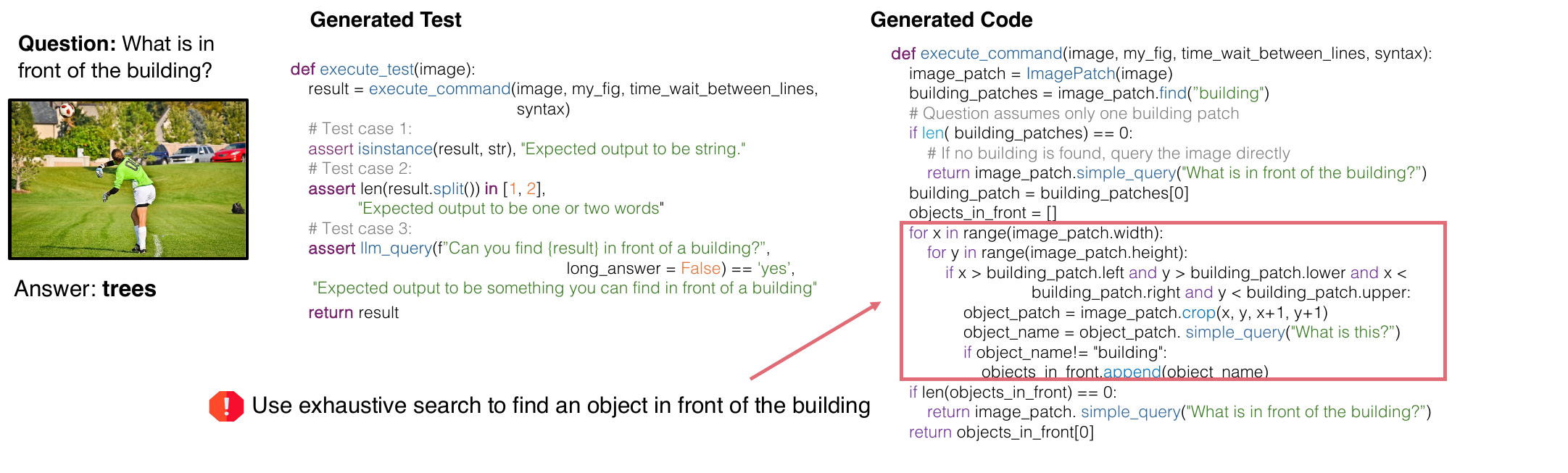}
    \caption{Example of inefficient code generated by Llama3-70b.  
    }
    \label{fig:appendix_llama70b_Failure}
\end{figure*}

Moreover, we identified a behavior unique to Llama3-70B, which tends to generate code with high time complexity. As illustrated in Fig.~\ref{fig:appendix_llama70b_Failure}, Llama3-70B often employs an exhaustive search to locate an object, even when a more efficient method like \texttt{find()} could be used. To handle these cases, we implemented a timer to raise an error if the execution exceeds 3 minutes, categorizing such instances as errors.

\section{Generated Property Test Case Analysis}
\label{sec:appendix_genPropTest}

\begin{table}[t!]
\centering
\small
\resizebox{\columnwidth}{!}{
\begin{tabular}{llcc}
\toprule
\textbf{Method} & \textbf{Dataset} & \textbf{Acc.} & \textbf{Toxic rate} \\
\midrule
Visual Grounding & RefCOCO         & 89.0\%         & 0.02\% \\
Visual Grounding & RefCOCO+        & 84.8\%         & 0.03\% \\
\bottomrule
\end{tabular}
}
\caption{Accuracy and toxic rate of generated property test cases on visual grounding tasks with Llama3-8B. APIs are utilized in visual grounding property test cases.}
\label{tab:appendix_testcase_eval}
\end{table}
First, \Cref{tab:appendix_testcase_eval} shows the evaluation of our generated visual grounding property test cases using the same two metrics as in~\Cref{tab:testcase_acc}. RefCOCO+ has lower accuracy and a higher toxic rate compared to RefCOCO, which can be due to the more complex queries within the RefCOCO+ dataset. 

\begin{figure}[t]
\centering
  \includegraphics[width=0.8\columnwidth]{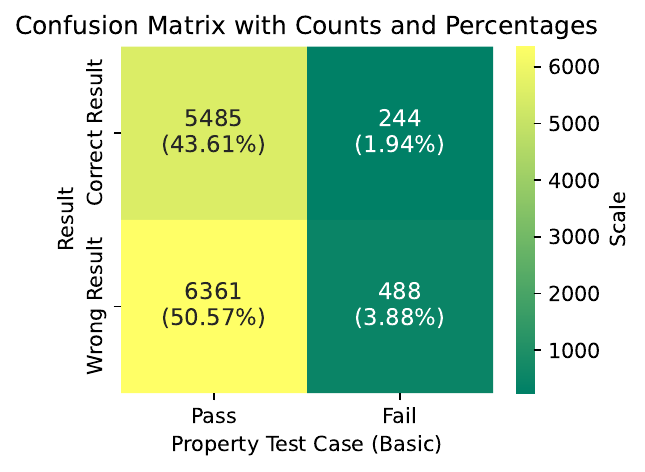}
  \caption{Confusion Matrix of the basic generated property test cases on GQA using Llama3-8B. We show the counts of correct and incorrect results, further divided by whether they passed or did not pass the generated property test case.}
  \label{fig:Basic_confusion_matrix}
\end{figure}
Additionally, we depict a confusion matrix of basic VQA property test cases on GQA using Llama3-8B in Fig.~\ref{fig:Basic_confusion_matrix}. The matrix depicts a high number of false positives because most basic VQA property tests check for data type, word length, and binary answers (yes or no), which can pass despite incorrect results. 

\begin{figure}[t!] 
    \centering
    \begin{subfigure}[b]{0.48\columnwidth}
        \centering
        \includegraphics[width=\textwidth]{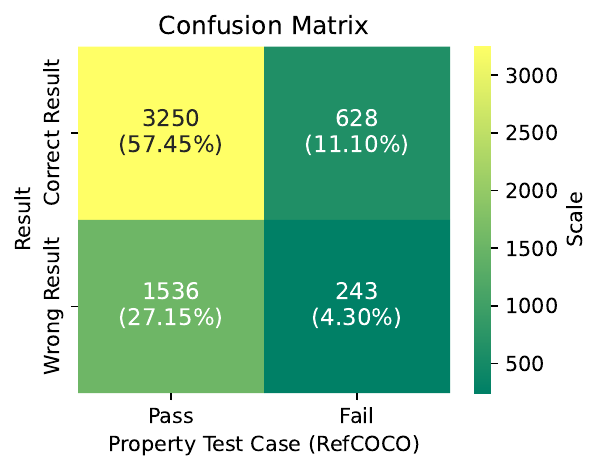}
        \caption{RefCOCO}
        \label{fig:refcoco_confusion_matrix}
    \end{subfigure}
    \hfill
    \begin{subfigure}[b]{0.48\columnwidth}
        \centering
        \includegraphics[width=\textwidth]{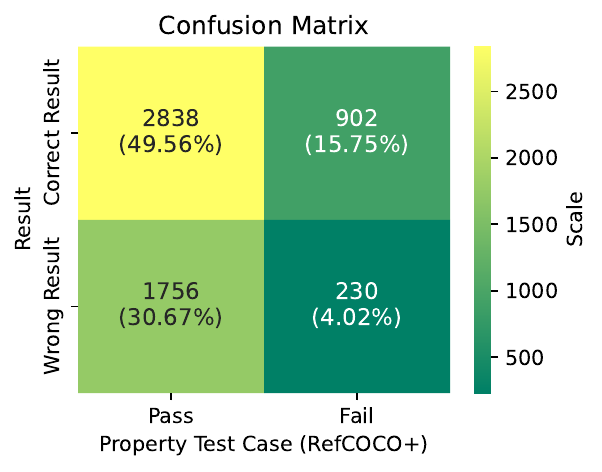}
        \caption{RefCOCO+}
        \label{fig:refcocoplus_confusion_matrix}
    \end{subfigure}
    \caption{Confusion Matrix for visual grounding property test cases on RefCOCO and RefCOCO+ using Llama3-8B. We consider the result to be correct if the IoU exceeds a threshold of 0.7.}
    \label{fig:appendix_Confusion_Matrix_RefCOCO}
\end{figure}
Fig.~\ref{fig:appendix_Confusion_Matrix_RefCOCO} plots the confusion matrix for visual grounding property test cases on RefCOCO and RefCOCO+. Half of the dataset falls under true positives (57.5\% on RefCOCO and 50.0\% on RefCOCO+), with a low true negative rate (0.04\% on RefCOCO and RefCOCO+), indicating the high quality of our generated property test cases. We observe a high number of false positives, similar to other datasets. This may be due to instances where, even if the IoU is below the threshold of 0.7, there is still an object or property that matches the query.


\end{document}